\documentclass[10pt, a4paper]{article}

\usepackage{arabtex}
\usepackage{utf8}
\setcode{utf8}

\DeclareFontFamily{U}{xnsh}{}%

\DeclareFontShape{U}{xnsh}{m}{n}{%
   <-6> sfixed * [6.0] xnsh14
      <6-10> s * [1.20] xnsh14
         <10><10.95><12><14.4><17.28><20.74><24.88> s * [1.0] xnsh14
         }{}

\DeclareFontShape{U}{xnsh}{bx}{n}{%
   <-6> sfixed * [6.0] xnsh14bf
   <6-10> s * [1.20] xnsh14bf
   <10><10.95><12><14.4><17.28><20.74><24.88> s * [1.0] xnsh14bf
}{}

\usepackage{lrec-coling2024} 
\usepackage{algorithm}
\usepackage{algpseudocode}
\usepackage{textcomp}
\usepackage{tcolorbox}
\usepackage{xcolor}

\definecolor{darkgreen}{HTML}{006400}

\title{\textbf{UQA: Corpus for Urdu Question Answering}}

\name{Samee Arif, Sualeha Farid, Awais Athar, Agha Ali Raza} 

\address{LUMS, LUMS, EMBL-EBI, LUMS \\
         Lahore, Pakistan; Lahore, Pakistan; Hinxton, UK; Lahore, Pakistan \\
         23100088@lums.edu.pk, 23100133@lums.edu.pk, awais@ebi.ac.uk, agha.ali.raza@lums.edu.pk\\}

\abstract{
This paper introduces UQA, a novel dataset for question answering and text comprehension in Urdu, a low-resource language with over 70 million native speakers. UQA is generated by translating the Stanford Question Answering Dataset (SQuAD2.0), a large-scale English QA dataset, using a technique called EATS (Enclose to Anchor, Translate, Seek), which preserves the answer spans in the translated context paragraphs. The paper describes the process of selecting and evaluating the best translation model among two candidates: Google Translator and Seamless M4T. The paper also benchmarks several state-of-the-art multilingual QA models on UQA, including mBERT, XLM-RoBERTa, and mT5, and reports promising results. For XLM-RoBERTa-XL, we have an \textbf{F1 score of 85.99 and 74.56 EM}. UQA is a valuable resource for developing and testing multilingual NLP systems for Urdu and for enhancing the cross-lingual transferability of existing models. Further, the paper demonstrates the effectiveness of EATS for creating high-quality datasets for other languages and domains. The UQA dataset and the code are publicly available at \url{www.github.com/sameearif/UQA}.
 \\ \newline  }

\begin{document}

\maketitleabstract

\section{Introduction}
The growth of natural language processing (NLP) tasks and datasets in English has been remarkable. However, expanding the reach of NLP to languages other than English, especially those that are lower on digital resources, is crucial for advancing multilingual AI systems. Among such languages, Urdu, with over 70 million native speakers\footnote{\url{www.britannica.com/topic/Urdu-language}}, stands as a significant yet underrepresented language in the NLP domain.

The Stanford Question Answering Dataset 2.0 (SQuAD2.0) \citep{rajpurkar2018know} is a benchmark for evaluating machine comprehension of text, but it is limited to English-based systems. There are two categories of questions: (1) Answerable questions: These are questions for which a clear, definite answer can be extracted directly from the provided passage or context (2) Unanswerable questions: These are questions for which the answer cannot be found in the provided passage but they look similar to answerable questions. Figure 1 shows examples of answerable and unanswerable question from SQuAD2.0.

Translating SQuAD2.0 into other languages seems like a straightforward task, but it comes with its own set of challenges, mainly when the job requires mapping the start index of the answer in the English context to the start index in the Urdu context. The introduction of the "Enclose to Anchor, Translate, Seek" (EATS) technique addresses this very challenge by enclosing the answer within a context using a specific delimiter, translating the enclosed context, and then seeking the delimiter's position post-translation.

In this study, we contrast the outputs of popular translation models, including Google Translator\footnote{\label{google}\url{cloud.google.com/translate/docs/reference/rest}} and Seamless M4T \citep{barrault2023seamlessm4t}. Through rigorous evaluation, we measure inter-rater agreement using Krippendorff's alpha \citep{krippendorff2004content} to discern the most consistent and reliable translation method among the contenders. The selected model then serves as our primary tool in the EATS technique to produce the Urdu-translated dataset.

We intend for our work to serve as a tool to further the development of Urdu NLP tools to enable access to mainstream language applications among Urdu speakers. Due to the dataset's large size and high quality, it can serve as a valuable resource to train LLMs in Urdu and create domain-specific applications to empower under-served populations via educational and health resources.

\begin{figure}[!ht]
    \begin{center}
        \begin{tcolorbox}        
        \textbf{Paragraph:}\\The further decline of Byzantine state-of-affairs paved the road to a third attack in \textcolor{darkgreen}{1185}, when a large Norman army invaded Dyrrachium, owing to the betrayal of high Byzantine officials. Some time later, Dyrrachium—one of the most important naval bases of the Adriatic—fell again to Byzantine hands.
        \\ \\ \textbf{Answerable Question:}\\When did the Normans attack Dyrrachium?
        \\ \textbf{Answer:} 1185
        \\ \\ \textbf{Unanswerable Question:}\\Who betrayed the Normans?        
        \end{tcolorbox}
        \caption{Question types example from SQuAD2.0}
    \end{center}
\end{figure}


\section{Related Work}
A large number of datasets for question-answering and text comprehension systems have been created for English. WikiQA \citep{yang-etal-2015-wikiqa} was introduced in 2015 - it included 3,047 questions and 29,258 sentences, where 1,473 sentences were labeled as answer sentences to the questions. Soon after, \citealp{rajpurkar2016squad} introduced the Stanford Question Answering Dataset (SQuAD) created by crowdworkers posing questions on Wikipedia articles. They compiled 100,000+ questions for the task of machine comprehension of text. In an attempt to create more robust question-answering systems, a more challenging dataset titled SQuAD2.0 \citep{rajpurkar2018know} was then introduced by expanding on the work done for SQuAD \citep{rajpurkar2016squad} - this introduced 50,000 unanswerable questions on top of the original dataset written adversarially by crowdworkers to look similar to answerable ones. Other corpora including HotpotQA \citep{yang2018hotpotqa} containing 113k Wikipedia-based question-answer pairs, TriviaQA \citep{joshi2017triviaqa} with over 650K question-answer-evidence triples, and Meta’s bAbI tasks data \citep{DBLP:journals/corr/WestonBCM15} were created to introduce greater complexity in data to train more capable QA systems.

Datasets for training cross-lingual functionality in QA systems were introduced in a multilingual context. The MLQA dataset was introduced by \citealp{lewis2019mlqa}. It contains QA instances in SQuAD format in seven languages (English, Arabic, German, Spanish, Hindi, Vietnamese, and Simplified Chinese) and was built using an alignment strategy on Wikipedia articles. They generated over 12,000 instances in English and 5,000 instances in each other language. Similarly, XQuAD \citep{artetxe2019cross} (13,000 examples spanning 11 languages), XQA presented by \citealp{liu-etal-2019-xqa} (28,000 instances in 9 languages), TyDi by \citealp{clark-etal-2020-tydi} (204,000 examples in 11 languages), Xor-QA \citep{asai-etal-2021-xor} (40,000 instances across seven languages), and MKQA \citep{longpre2021mkqa} (260,000 examples in 26 languages) was introduced for multilingual question answering systems.

There has been comparatively less work for monolingual non-English corpora - particularly for low-resource languages. A popular method of generating resources for such languages has been the translation of datasets for English into the target language employing different machine translation implementations. Some examples of such work are ParSQuAD \citep{abadani2021parsquad} for Persian, SQuAD-it \citep{10.1007/978-3-030-03840-3_29} for Italian, Vietnamese SQuAD\footnote{\url{www.kaggle.com/datasets/nkhachao/vietnamese-squad}}, K-QuAD\footnote{\url{www.github.com/Di-lab-Yonsei/K-QuAD}} for Korean, and Arabic-SQuAD \citep{mozannar-etal-2019-neural}.

For Urdu, question-answering resources are scarce. Some datasets have been presented, such as UQuAD\footnote{\url{www.github.com/ahsanfarooqui/UQuAD---Urdu-Question-Answer-Dataset/tree/main}} containing 499 questions and 27 paragraphs. Urdu Open-Ended Question Answer Text Dataset\footnote{\url{www.futurebeeai.com/dataset/prompt-response-dataset/urdu-open-ended-question-answer-text-dataset}} with 5000+ question-answer pairs, and Urdu Closed-Ended Question Answer Text Dataset\footnote{\url{www.futurebeeai.com/dataset/prompt-response-dataset/urdu-closed-ended-question-answer-text-dataset}} also containing 5,000+ question-answer pairs are both human-generated datasets however they are not open source and do not provide any metrics or comments regarding the quality of the data. UQuAD1.0 \citep{kazi2021uquad1} is a work involving the translation of SQuAD1.0 \citep{rajpurkar2016squad} containing 49,000 question-answer pairs from which 45,000 are translated from SQuAD \citep{rajpurkar2016squad} (53\% of the data, the remaining 47\% was discarded) and 4,000 were manually generated via crowdsourcing. 
However, the dataset is not publicly available. Therefore, to the best of our knowledge, no large, high-quality, publicly available dataset exists for Urdu question answering and text comprehension - making our contribution a valuable and important step towards developing tools for this low-resource language.

\section{Methodology}
Translating the Stanford Question Answering Dataset (SQuAD2.0) \citep{rajpurkar2018know} into Urdu presents a unique set of challenges, with one of the foremost difficulties lying in accurately identifying the answer's starting position within the translated context paragraph. This challenge stems from the linguistic differences between the source and target language. Both languages have different grammatical structures, vocabulary, and idiomatic expressions, meaning there is no one-to-one mapping between the words in the source text and the translated text. The source language and the target language also have different word order and sentence structure, that is, English follows subject-verb-object (SVO) order, and Urdu follows subject-object-verb (SOV) order. Therefore, addressing this challenge requires a robust method for aligning and matching the answer spans.

\subsection{Translation Model Selection}
All SQuAD2.0 \citep{rajpurkar2018know} context paragraphs were split into sentences using the python NLTK \citep{bird-loper-2004-nltk} sentence tokenizer for our experiments. For experiment 1, we selected a set of 100 sentences to conduct a pilot test. This smaller subset allowed us to assess the viability of our translation methodology and the overall experimental design in a controlled, manageable environment. In experiment 2 we selected a set of 1,512 sentences from a total of 100,026 sentences. The minimum required sample size for a confidence level of 99\% with a 3\% margin of error was calculated to be 1,030 using the population size (100,000) - we therefore took a sample of 1,512 sentences. The selected sentences for both experiments were subsequently passed through two machine translation systems: Google Translator\textsuperscript{\ref{google}} and Facebook Seamless M4T \citep{barrault2023seamlessm4t}.

In evaluation 1, three annotators (computer science researchers - native Urdu speakers) were presented with two anonymized machine translation systems, one of which was the Seamless M4T model and the other was Google Translator. The annotators were each assigned a total of 100 identical sentences. Their task involved labeling the data to indicate one of the following: (1) both translators produced the same output quality for a sentence, (2) Seamless M4T provided a better translation, or (3) Google Translator provided a better translation. The Google Translator was picked 14.33\% of the times, and Seamless M4T was picked 51.67\% of times, and both were considered to be of the same quality 34.0\% of times. To determine the inter-rater reliability, the Krippendorff’s alpha value was calculated and found to be 0.688, which, according to Krippendorff's interpretation \citep{krippendorff2004content} is sufficient for a tentative conclusions to be drawn.

In evaluation 2, twelve voters - undergraduate students, native Urdu speakers with English as medium of instruction - were asked to pick between the two translation models. The voters were given the same task as the annotators in Experiment 1. The Google Translator was picked 37.43\% of the times, and Seamless M4T was picked 54.37\% of the times, and both were considered to be of the same quality 8.20\% of times.

In summary, Seamless M4T consistently demonstrated superior translation quality in both evaluations when compared with Google Translator.

\subsection{Initial Experiments}
Our initial approach involved translating the SQuAD2.0 dataset \citep{rajpurkar2018know} by translating the question and the answer string and then translating the context sentence by sentence. When we reached the line where the answer string was present, we used string matching to find the translated answer. However, we encountered significant challenges: (1) The answer string in the context often underwent grammatical modifications when included within a paragraph; (2) The sentence tokenization libraries failed to detect all the abbreviations, resulting in low-quality sentence segmentation and, therefore, degraded translation results. As a result, relying solely on the exact string-matching approach proved to be insufficient for pinpointing the answer's start index.

In our second approach, we opted to translate the each undivided paragraph as a single unit (to retain the context and semantic meaning) instead of translating line by line. This shift in strategy allowed us to pass the translated paragraph alongside the translated question and answer to a Large Language Model (LLM), such as LLaMA 2 \citep{touvron2023llama} and GPT-3.5 \citep{brown2020language}, in an attempt to determine the answer's start and end positions automatically. However, the models performed poorly on our text as they did not predict the correct start and end points in the Urdu paragraphs.

Subsequently, we transitioned to using GPT-4 \citep{openai2023gpt4}, demonstrating promising results in accurately identifying answer positions within the translated context paragraph. However, the drawback of this approach was the significant computational cost associated with GPT-4, which rendered it impractical for this task. As a result, we had to reconsider our methodology to balance performance with computational efficiency and cost.

To address the challenges encountered in translating the SQuAD2.0 dataset into Urdu and accurately identifying answer start positions, we implemented a three-step solution illustrated in Figure 2.

\begin{figure}[!ht]
\begin{center}
\includegraphics[scale=0.6]{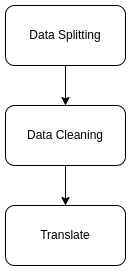} 
\caption{Three-step solution}
\label{fig.2}
\end{center}
\end{figure}

\subsection{Implementation}
\subsubsection{Data Splitting}
When a large paragraph (defined as text containing more than 1,000 characters) is passed through Seamless M4T \citep{barrault2023seamlessm4t}, it tends to summarize or drop the last few sentences. To illustrate this behavior, Table 1 presents sentences from a paragraph of length 1,427 characters. The entire paragraph is passed through the translator, then the English paragraph is manually split into sentences, and the corresponding sentences are extracted from the translated paragraph. It is evident from the output that the last sentence is not fully translated.

To address this issue, we initially identified paragraphs with a length equal to or exceeding 1,000 characters. We then manually divided these 3,307 paragraphs into smaller segments, ensuring that each paragraph segment had a length of less than 1,000 characters.

\renewcommand{\arraystretch}{1.5}
\begin{table}[!ht]
\begin{center}
\centerline{\includegraphics[width=1\columnwidth]{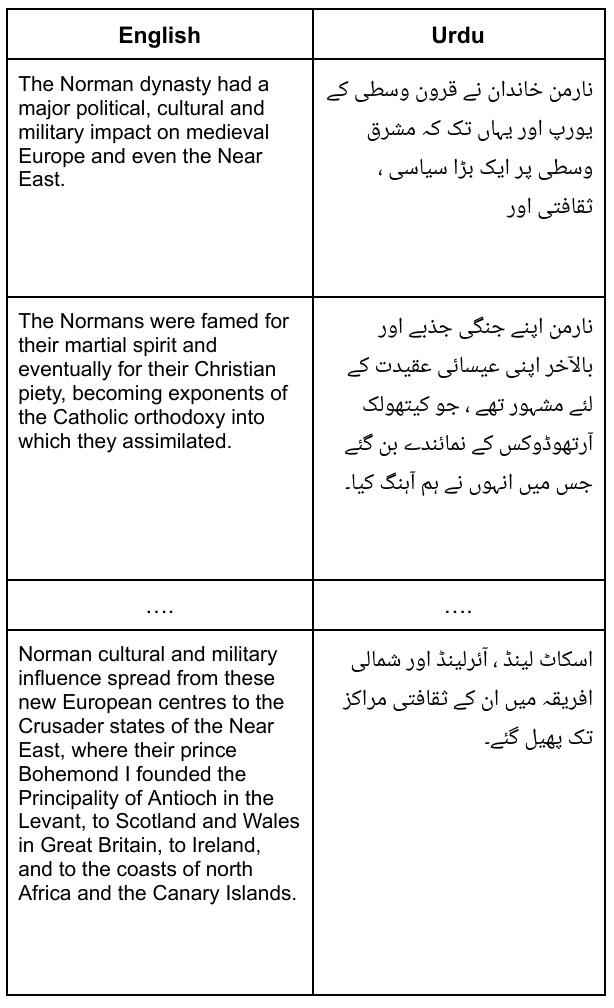}}
\caption{\small{English Paragraph to Urdu translation}}
\label{fig:pos-struct}
\end{center}
\vskip -0.3in
\end{table}

\subsubsection{Data Cleaning - EATS}
To ensure that answer strings are retained in the translated text and unaffected by any text misalignment during the translation process, we introduce the EATS technique: Enclose to Anchor, Translate, Seek. The process involves first highlighting the answer string in the original text by enclosing it in delimiters, then passing the text through a translator and seeking the answers in the target language by looking for the delimiters. Thus, the first part of our process was to ensure that the data was in the following format:

\begin{center}
    \textit{Infrared radiation is used in industrial, $••$scientific$••$ and medical applications.}
\end{center}

The answer was marked with the delimiters because the removal of certain characters from the string would offset the answer start position, so '$••$' acts as a marker for the answer string in the context paragraph.

Seamless M4T \citep{barrault2023seamlessm4t} sometimes fails to handle semicolons '$;$', en dashes '$–$' that are used between figures to represent the range and em dashes '$—$' that are used to create a strong break in a sentence, emphasizing an interruption or additional information. To account for this, we replaced all the semicolons with the Arabic semicolon '\RL{؛}' and  en dashes and em dashes with double hyphens '$--$' before the translation process started. Following this, all double quotation marks were removed from the text to ensure that only the answer string is enclosed within the specific double quotation mark ‘\textquotedbl' (i.e. U+0022 in UTF-16 encoding) for the translation process. In the final step '$••$' was replaced with ‘\textquotedbl'. This data cleaning procedure was carried out for all the paragraphs, questions, and answers in SQuAD2.0 \citep{rajpurkar2018know}.

\subsubsection{Translation}
As we had used quotation marks to highlight the answer paragraphs, they would serve as essential markers to identify the answer's start and end positions within the translated paragraph and precisely locate the answer within the text. Algorithm 1 outlines the pseudo-code for this process. In the algorithm, variable $paragraphs$ is a list containing either a single entity if it contains less than or equal to 1000 characters, otherwise it contains multiple entities i.e. sub-paragraphs - this is due to the splitting methodology defined in section 3.3.1. Following the translation, the double hyphens '$--$' in between digits were replaced with en dash '$–$', and the rest (i.e those not between digits) were replaced with em dash '$—$'. 

\begin{algorithm}
\begin{algorithmic}
\For{$([paragraphs],\ question,\ answer)$ \textbf{in} $data$}
    \State $question_t \leftarrow translate(question)$
    \State $answer_t \leftarrow translate(answer)$
    \State $paragraph_t \leftarrow [\ ]$
    
    \For{$para$ \textbf{in} $paragraphs$}
        \If{$'$••$'$ \textbf{is in} $para$}
            \State $Replace\ '$••$'\ with\ '$\textquotedbl$' \ in\ para$ 
            \State $para \leftarrow translate(para)$
            \If{$para.count('$\textquotedbl$')\ \neq \ 2$}
                \State \textbf{break}
            \EndIf
        \Else
            \State $para \leftarrow translate(para)$
        \EndIf
        \State $paragrapht_t.append(para)$
    \EndFor
\EndFor
\end{algorithmic}
\caption{Translation Algorithm}
\end{algorithm}

    
            
    

Our methodology failed to retain quotation marks for only 392 out of 11,858 questions in the dev set and 5,574 out of 130,319 questions in the train set. Therefore, from a total of 142,177 questions only 5,966 were discarded which highlights the effectiveness of our approach, demonstrating a high degree of precision. In addressing the issue of missing quotation marks in a minor subset of our dataset, we found that GPT-4 \citep{openai2023gpt4} could effectively correct these errors. However, considering that the erroneous data was only 4.2\% of the overall dataset, we ultimately decided against correcting this subset using GPT-4 with the aim of minimizing expenses.

Following the translation and implementation of EATS, we generated a total of 124,745 questions in the train set and 11,466 questions in the dev set. A breakdown of the number of questions in each category is provided in Table 2.

\newcolumntype{B}{>{\centering\arraybackslash}p{1.10cm}}
\newcolumntype{C}{>{\centering\arraybackslash}p{1.10cm}}
\renewcommand{\arraystretch}{1.5}

\begin{table}[!ht]
    \begin{center}    
    \begin{tabularx}{\columnwidth}{|X|B|C|}
        \hline
        \textbf{} & \textbf{Dev} & \textbf{Train}\\
        \hline

        \textbf{Answerable Questions} & 5,811 & 83,018\\
        \hline

        \textbf{Unanswerable Questions} & 5,655 & 41,727\\
        \hline
        
    \end{tabularx}
    \caption{Dataset summary}
    \end{center}
\end{table}

\section{Evaluation and Discussion}

We fine-tuned and evaluated different variants of mBERT \citep{devlin-etal-2019-bert}, XLM-RoBERTa (abbreviated as XLM-R)  \citep{conneau-etal-2020-unsupervised, goyal2021largerscale} and mT5 \citep{xue2021mt5} on our dataset. All the models were fine-tuned for 4 epochs, learning rate for XLM-R models and mBERT was set to $2e^{-5}$ and $5e^{-5}$ for mT5 models. We used only answerable questions for fine-tuning on the train set and evaluating these models on the dev set. The performance of the models is quantified using two common metrics: Exact Match (EM) and F1 Score using the Huggingface wrapper\footnote{\url{www.huggingface.co/spaces/evaluate-metric/squad}} for the official SQuAD evaluation script by \citealp{rajpurkar2016squad}. All the models were trained for six epochs and the best checkpoint of each model was evaluated on the dev part of the dataset. Table 3 summarizes the results of the experiments.

\newcolumntype{B}{>{\centering\arraybackslash}p{1.35cm}}
\newcolumntype{C}{>{\centering\arraybackslash}p{1.35cm}}
\renewcommand{\arraystretch}{1.5}
\begin{table}[!ht]
    \begin{center}    
    \begin{tabularx}{\columnwidth}{|X|B|C|}
        \hline
        \textbf{Model} & \textbf{F1 Score} & \textbf{Exact Match (EM)}\\
        \hline

        mBERT & 64.72 & 45.50\\
        \hline

        mT5-Small & 67.24 & 52.37\\
        \hline

        mT5-Large & 84.20 & 71.26\\
        \hline

        XLM-R & 78.00 & 65.67\\
        \hline

        XLM-R-Large & 84.42 & 72.24\\
        \hline

        \textbf{XLM-R-XL} & \textbf{85.99} & \textbf{74.56}\\
        \hline
        
    \end{tabularx}
    \caption{Evaluation summary}
    \end{center}
\end{table}

We can see that the XLM-R-XL performs the best for both metrics, with mT5-Large closely following. This can be explained by the number of parameters (3.5B vs 1.2B) as well as differences in the size and quality of their original training corpora.

Comparing the results with existing state-of-the-art models for Urdu and similar languages (Persian and Arabic), Table 4 shows that our XLM-R-XL UQA model outperforms the best reported scores. While these results are not directly comparable with existing work due to differences in model parameter sizes, results of our evaluation on comparable models including mBERT and XLM-R (table 3) show that models trained on UQA outperform those presented for Arabic-SQuAD (BERT), and UQuAD1.0 (XLM-R). This improvement can be attributed to the quality of translation as well as the size of our training data. 

Incorporating unanswerable questions from our dataset into the training set could present a valuable opportunity to enhance model performance. Training on both answerable and unanswerable questions might empower the model to better discern between the two, potentially refining its ability to identify and respond to answerable queries with increased precision.

\newcolumntype{B}{>{\centering\arraybackslash}p{1.7cm}}
\newcolumntype{C}{>{\centering\arraybackslash}p{0.90cm}}
\begin{table}[!ht]
    \begin{center}    
    \begin{tabularx}{\columnwidth}{|X|B|C|C|}
        \hline
        \textbf{Dataset} & \textbf{Model} & \textbf{F1 Score} & \textbf{Exact Match (EM)}\\
        \hline

        ParSQuAD & ALBERT  & 70.84 & 67.73\\
        \hline

        Arabic-SQuAD & BERT & 61.30 & 34.20\\
        \hline

        UQuAD1.0  & XLM-R & 66.00 & 36.00\\
        \hline

        \textbf{UQA} & \textbf{XLM-R-XL} & \textbf{85.99} & \textbf{74.56}\\
        \hline
        
    \end{tabularx}
    \caption{Comparison with existing models}
    \end{center}
\end{table}

\section{Conclusion} 
In this paper, we present the process of creating a question answering corpus for Urdu and make UQA publicly available. By training multiple state of the art question answering models on our datasets to get promising evaluation scores, we demonstrate the suitability of our dataset for training and evaluation of transformer based models. Future work can include building on to the dataset with domain specific data to fine-tune models - particularly LLMs - for a specific use case such as providing health care facilities to low resource areas.

In the translation process, we primarily relied on the selected model's inherent accuracy, given that no translation model guarantees 100\% accuracy. We also did an extensive evaluation of the translation models to ensure that the one with the highest accuracy was used. In a low-resource language like Urdu achieving perfect translation accuracy can be a challenge, the large size of the dataset also makes manual fixes infeasible.

This paper also forms the groundwork for a pipeline to produce further domain-specific QA resources for Urdu without the need for translation by relying directly on question generation models that can be trained on UQA.

Our work for resource generation in low resource languages, therefore, creates the opportunity to address the challenge of large-scale data generation required for language models across diverse languages and domains. Particularly in contexts where native data in the target language is sparse or unavailable.

\section{Acknowledgements}
We are grateful for the time and effort put in by our research interns who annotated our data: Aamina Jamal Khan, Abdullah Hashmat, Abeer Aamir, Daanish Uddin Khan, Eesha Kamran, Maham Ahmed Bhimra, Maryam Usman, Muhammad Faisal Nazir, Muhammad Khaquan, Muhammad Shayan, Omer Tafveez, and Qasim Anwar who supported our work.

\nocite{*}
\section{Bibliographical References}\label{sec:reference}

\bibliographystyle{lrec-coling2024-natbib}
\bibliography{lrec-coling2024-example}


\end{document}